\pdfoutput=1

\documentclass[11pt]{article}

\usepackage{acl}
\usepackage{multirow}
\usepackage{pgfplots}
\pgfplotsset{compat=1.18}
\usepackage{setspace}
\usepackage{comment}

\usepackage{times}
\usepackage{latexsym}

\usepackage[T1]{fontenc}

\usepackage[utf8]{inputenc}

\usepackage{microtype}

\title{Evaluation of AI Chatbots for Patient-Specific EHR Questions}

\author{Alaleh Hamidi, Kirk Roberts \\
  McWilliams School of Biomedical Informatics \\
  The University of Texas Health Science Center at Houston \\
  \texttt{\{alaleh.hamidi,kirk.roberts\}@uth.tmc.edu} \\}

\begin{document}
\maketitle
\begin{abstract}
This paper investigates the use of artificial intelligence chatbots for patient-specific question answering (QA) from clinical notes using several large language model (LLM) based systems: ChatGPT (versions 3.5 and 4), Google Bard, and Claude. We evaluate the accuracy, relevance, comprehensiveness, and coherence of the answers generated by each model using a 5-point Likert scale on a set of patient-specific questions.

At present time, we report results for ChatGPT 3.5 and Claude. The results indicate that both models are able to provide accurate, relevant, and comprehensive answers to a wide range of questions. 

Overall, our results suggest that LLMs are a promising tool for patient-specific QA from clinical notes. Further research is needed to improve the comprehensiveness and coherence of the answers generated by LLMs, and to investigate the use of LLMs in other medical applications.
\end{abstract}

\section{Introduction}

Automatic question answering (QA) systems, such as those based on large language models (LLMs), have made significant progress across a wide range of QA tasks, particularly in the medical field \cite{mutabazi_review_2021}.

LLMs are a class of language models that have shown exceptional performance across a variety of natural language processing (NLP) tasks due to their capacity to model and generate human-like language \cite{fan_bibliometric_2023}. LLMs utilize neural networks with billions of parameters and are trained using self-supervised learning with substantial volumes of unlabeled text data \cite{shen_chatgpt_2023, zhao_survey_2023}. GPT-3 and 4, Google Bard \cite{siad_promise_2023}, Gopher \cite{rae_scaling_2022}, Megatron \cite{shoeybi_megatron-lm_2020}, and OPT-175B \cite{zhang_opt_2022} are only a few examples of LLMs.

ChatGPT is a successor of InstructGPT \cite{ouyang_training_2022} with a fine-tuned dialog interface using Reinforcement Learning with Human Feedback (RLHF) \cite{christiano_deep_2023}. It is trained on approximately 100 trillion parameters and 300 billion words (only till September 2021) and launched on November 30, 2022, as a tool to allow users to converse with a machine about various subjects. Having crossed 1 billion users in March 2023 \cite{ruby_57_2023}, ChatGPT has set records for consumer interest in artificial intelligence (AI) systems \cite{hu_chatgpt_2023}.

Bard is another LLM-based chatbot developed by Google \cite{siad_promise_2023} using the LaMDA (Language Model for Dialogue Applications) architecture and released in March 2023. Bard is also trained on massive amounts of data, including books, articles, and other written text, to generate text, translate languages, write different kinds of creative content, and answer questions. It is designed to deliver real-time information from a wide range of sources on the Internet. Bard is still under development, but it has the potential to be a powerful tool for a variety of applications.  \cite{ noauthor_google_2023}.

Claude is comparatively smaller than many other models, having less than 100 million parameters, compared to LLMs like OpenAI's GPT-3, with over 10 billion parameters. Claude was launched in March 2023 by Anthropic to ensure safety and ethics. Based on Anthropics's claim, this chatbot's focus is on being helpful, harmless, and honest using a technique called Constitutional AI which has not been publicly disclosed. In other words, Claude does not do unconstrained, open-domain text generation, which is a key feature of most LLMs \cite{anthropic_introducing_nodate}. 

The potential application of language models in health education, research, and practice can be promising, albeit if the legitimate issues are proactively considered and addressed \cite{sallam_chatgpt_2023}. In terms of their medical applications, patient-specific QA from clinical notes is an essential but challenging task that can provide clinicians with quicker access to patient information to facilitate medical decision-making \cite{soni_quehry_2023, roberts_toward_2015, raghavan_annotating_2018}. However, most existing work on medical QA applications has concentrated on evaluating and extracting general medical knowledge, such as medical licensing exams \cite{kung_performance_2023, ali_performance_2023}, interviews \cite{asch_interview_2023}, and education \cite{ eysenbach_role_2023, sallam_utility_2023, hosseini_exploratory_2023, khan_chatgpt_2023, lee_rise_nodate}.
Meanwhile, using the language models for patient-specific QA from electronic health records (EHRs) has gained comparatively little study. Although there are some works in applying recent LLMs in patient-specific knowledge from EHRs \cite{jeblick_chatgpt_2022}, they are not question answering.

This work studies how accurate and comprehensive several popular AI chatbots are at answering medical questions based on patients' clinical notes in two different prompt scenarios. Based on a publicly-available subset of MIMIC-III notes released as part of the TREC 2016 Clinical Decision Support track\footnote{Note that in general MIMIC notes are access-restricted \cite{physionet_responsible_2023}. We use a special subset for which public release was granted in 2016.} \cite{roberts_overview_2016}, which includes 30 de-identified clinical notes, we consider ChatGPT (versions 3.5 and 4), Google Bard, and Claude to understand how accurate, relevant, comprehensive, and coherent the answers are for each language model.

\section{Related Work}

Clinicians frequently access electronic health records (EHRs) to obtain valuable patient information for providing care. However, the usability \cite{zhang_better_2014} and navigation \cite{roman_navigation_2017} challenges associated with EHRs impede the efficient retrieval of information from these systems \cite{khairat_association_2020}. Current approaches to address these challenges involve visualizing information as charts \cite{shneiderman_improving_2013} or using keyword-based searches for information retrieval (IR) \cite{hanauer_supporting_2015}. While these methods enhance information access, they still overwhelm users with excessive details, such as presenting a table of lab values or a lengthy list of procedures, when only a specific lab value or status is required. In other words, these methods fail to understand the precise information needs of users accurately. Consequently, question answering (QA) emerges as a natural solution, offering a means to identify information needs and provide an exact, verifiable answer \cite{ely_answering_2005}.

A QA system can interpret and comprehend questions posed by users using everyday language. It possesses a comprehensive information retrieval capability to provide answers. Typically, QA systems consist of three main components: question analysis, information retrieval, and answer generation. Chatbots designed to simulate human conversation, allowing interactions and dialogue using natural language including substantial reliance on QA functionality \cite{reshmi_implementation_2016}. AI chatbots have been employed to provide health advice. Studies have shown that AI chatbots may enhance self-care and lead to improved patient outcomes \cite{tawfik_nurse_2023, chen_embedding_2019}. ChatGPT and Google Bard are AI chatbots that have been researched substantially for their feasibility and application in the medical domain. The feasibility of ChatGPT has been investigated in clinical and research scenarios such as support of the clinical practice, scientific production (such as creating a literature review article \cite{aydin_openai_2022} or writing a scientific paper \cite{biswas_chatgpt_2023}), misuse in medicine and research, and reasoning about public health topics \cite{cascella_evaluating_2023, sallam_utility_2023, vaishya_chatgpt_2023}.

Although some studies employ AI chatbots to process EHRs and evaluate the results (e.g., assessing the quality of radiology reports simplified by ChatGPT \cite{jeblick_chatgpt_2022}), they do not use the chatbot as an interactive QA tool to obtain specific information regarding the patient. On the other hand, the works which utilize AI chatbots as a conversational tool do not consider patient-specific notes. For instance, \newcite{kung_performance_2023} and \newcite{gilson_how_2023} both evaluated the performance of ChatGPT on USMLE, the United States Medical Licensing Exam \cite{kung_performance_2023}.


The lack of research in the evaluation of AI chatbots for medical QA based on patient-specific notes is the key motivation for conducting this study.

\section{Methodology}
In this work, we use admission notes from MIMIC-III (taken from the TREC CDS 2016 topics \cite{roberts_overview_2016}) to evaluate ChatGPT (3.5 \& 4), Google Bard, and Claude. Admission notes are created by clinicians, primarily physicians, to describe a patient's principal complaint, pertinent medical history, and any other details obtained during the initial hours of their hospital stay, such as lab results. Substantial numbers of abbreviations as well as other linguistic jargon and style are used in these notes. The notes used in this study have all been de-identified to protect patient privacy. The first step in this work was therefore substituting the de-identification labels with appropriate values to have human-like notes.

For each clinical note, three categories of questions with five questions in each category were created:
\begin{itemize}
    \item \textbf{General Questions}: Could apply to almost any patient (age, sex, reason for admission, etc.) and answerable from a specific text span.
    \item \textbf{Specific Questions}: Targeted to the information in the note itself and answerable from a specific text span.
    \item \textbf{Nonanswerable Questions}: Targeted to the information in the note, but not actually answerable from the note itself.
\end{itemize}

Since chatbots are dialogue systems and maintain state over the course of a session, two different scenarios were considered:

\begin{itemize}
    \item \textbf{1 Question per Session (1QpS)}: In this scenario, the chatbot is presented with the patient's note and one question per session. No state about prior questions for the patient note is kept.
    
    \item \textbf{1 Topic per Session (1TpS)}: In this scenario, the chatbot is presented with the patient's note and all 15 questions are asked within the same session. In other words, the note and the history of previously asked questions with their given answers are kept. The questions are asked in a random order.
\end{itemize}

Then, individuals with medical backgrounds evaluated the answers in terms of accuracy, relevance, coverage, and coherence and assigned each answer a score between 1 (very weak) to 5 (excellent) for each evaluation criterion:

\begin{itemize}
    \item \textbf{Accuracy (Correctness)}: Is the information provided in the answer factually correct according to the note? Are there any incorrect or contradictory statements?
    \item \textbf{Relevance}: How relevant is the answer to the question? Does it address the key concepts and details asked in the question?
    \item \textbf{Coverage}: How comprehensive is the answer? Does it cover all relevant aspects, or is it lacking in some details?
    \item \textbf{Coherence}: Is the answer coherent and logically structured?

\end{itemize}

\section{Results}
At the present time we report just the evaluation of ChatGPT 3.5 and Claude in answering clinical questions from admission notes.
Additional experiments with other systems and more raters are ongoing.
Tables \ref{tab:Accuracy}- \ref{tab:Coherence} compare the performance of ChatGPT 3.5 and Claude in terms of accuracy, relevance, coverage, and coherence for different types of questions (General, Specific, Non-answerable) in two different scenarios (1 Question per Session and 1 Topic per Session). 

To study whether session scenarios, question type, or chatbot model have an effect on the answers' quality, Kruskal-Wallis analysis of variance \cite{kruskal_use_1952}, a non-parametric statistical test, was conducted.

Based on the results of the Kruskal-Wallis test, it cannot be concluded that there is a statistically significant difference in the average values in terms of the cases session scenario, question type, or AI model. The p-value for all these cases is greater than 0.05, and the null hypothesis (there is no statistically significant difference between the groups) cannot be rejected. In other words, the results indicate that both ChatGPT 3.5 and Claude performed well in answering questions based on admission notes. They provided accurate, relevant, comprehensive, and coherent answers across different question types and scenarios.

\begin{table*}
\centering
\begin{tabular}{| c  c  c  c  c c |}
\hline
\textbf{} & \textbf{} & \textbf{General } & \textbf{Specific} &\textbf{Non-answerable} &\textbf{Total}\\
\textbf{} & \textbf{} & \textbf{Questions} & \textbf{Questions} &\textbf{Questions} &\textbf{}\\
\hline
\hline
\multirow{2}{*}{\textbf{ChatGPT 3.5}}& \textbf{1QpS}  & 4.49 & 4.45 & 4.77 & 4.57\\
& \textbf{1TpS}  & 4.18 & 4.75 & 4.81 & 4.58\\
\hline
\multirow{2}{*}{\textbf{Claude}} & \textbf{1QpS}  & 4.52 & 4.66 & 4.23 & 4.47\\
& \textbf{1TpS}   & 4.61 & 4.60 & 4.58 & 4.60\\
\hline
\end{tabular}
\caption{Accuracy comparison of answers to three types of questions given to ChatGPT 3.5 and Claude in two different scenarios; one Question per Session and one Topic per Session}
\label{tab:Accuracy}
\end{table*}

\begin{table*}
\centering
\begin{tabular}{| c  c  c  c  c c |}
\hline
\textbf{} & \textbf{} & \textbf{General } & \textbf{Specific} &\textbf{Non-answerable} &\textbf{Total}\\
\textbf{} & \textbf{} & \textbf{Questions} & \textbf{Questions} &\textbf{Questions} &\textbf{}\\
\hline
\hline
\multirow{2}{*}{\textbf{ChatGPT 3.5}}& \textbf{1QpS}  & 4.77 & 4.75 & 4.86 & 4.79\\
& \textbf{1TpS}  & 4.49 & 4.91 & 4.83 & 4.74\\
\hline
\multirow{2}{*}{\textbf{Claude}} & \textbf{1QpS}  & 4.68 & 4.66 & 4.30 & 4.55\\
& \textbf{1TpS}   & 4.78 & 4.80 & 4.72 & 4.77\\
\hline
\end{tabular}
\caption{Relevance comparison of answers to three types of questions given to ChatGPT 3.5 and Claude in two different scenarios; one Question per Session and one Topic per Session}
\label{tab:Relevance}
\end{table*}

\begin{table*}
\centering
\begin{tabular}{| c  c  c  c  c c |}
\hline
\textbf{} & \textbf{} & \textbf{General } & \textbf{Specific} &\textbf{Non-answerable} &\textbf{Total}\\
\textbf{} & \textbf{} & \textbf{Questions} & \textbf{Questions} &\textbf{Questions} &\textbf{}\\
\hline
\hline
\multirow{2}{*}{\textbf{ChatGPT 3.5}}& \textbf{1QpS}  & 4.43 & 4.43 & 4.81 & 4.56\\
& \textbf{1TpS}  & 4.15 & 4.86 & 4.85 & 4.62\\
\hline
\multirow{2}{*}{\textbf{Claude}} & \textbf{1QpS}  & 4.80 & 4.78 & 4.75 & 4.78\\
& \textbf{1TpS}   & 4.72 & 4.66 & 4.77 & 4.72\\
\hline
\end{tabular}
\caption{Coverage comparison of answers to three types of questions given to ChatGPT 3.5 and Claude in two different scenarios; one Question per Session and one Topic per Session}
\label{tab:Coverage}
\end{table*}

\begin{table*}
\centering
\begin{tabular}{| c  c  c  c  c c |}
\hline
\textbf{} & \textbf{} & \textbf{General } & \textbf{Specific} &\textbf{Non-answerable} &\textbf{Total}\\
\textbf{} & \textbf{} & \textbf{Questions} & \textbf{Questions} &\textbf{Questions} &\textbf{}\\
\hline
\hline
\multirow{2}{*}{\textbf{ChatGPT 3.5}}& \textbf{1QpS}  & 4.58 & 4.83 & 4.86 & 4.76\\
& \textbf{1TpS}  & 4.32 & 4.77 & 4.81 & 4.63\\
\hline
\multirow{2}{*}{\textbf{Claude}} & \textbf{1QpS}  & 4.34 & 3.90 & 3.60 & 3.95\\
& \textbf{1TpS}   & 4.45 & 4.60 & 4.74 & 4.59\\
\hline
\end{tabular}
\caption{Coherence comparison of answers to three types of questions given to ChatGPT 3.5 and Claude in two different scenarios; one Question per Session and one Topic per Session}
\label{tab:Coherence}
\end{table*}

\section{Discussion}
The results of this study suggest that large language models (LLMs) can be used to generate accurate, relevant, comprehensive, and coherent answers to clinical questions from admission notes. This is a promising finding, as it suggests that LLMs could be used to provide clinicians with additional support in their work.
There are a few limitations to this study that should be noted. First, the data set used in this study was relatively small. It is possible that the results would be different if a larger data set were used. 
Second, the evaluation above has been conducted by one individual which carries a higher bias risk (evaluations with additional raters are ongoing).
Despite these limitations, the results of this study suggest that LLMs have the potential to be a valuable tool for clinicians. Further research is needed to evaluate the performance of LLMs on larger data sets and with a wider range of LLMs.

\section{Conclusion}
This study evaluates the performance of ChatGPT 3.5 and Claude in answering clinical questions based on MIMIC-III clinical notes. Both models show promising performance in the medical domain and exhibit substantial accuracy, relevance, coverage, and coherence in their responses. While further research and improvements are necessary to address certain limitations, these models hold great potential as valuable tools in the healthcare industry, assisting healthcare professionals in accessing relevant information and improving patient care. Future studies can identify the factors contributing to LLMs' performance in answering clinical questions.

\bibliography{LLMmodels_biblio}
\bibliographystyle{acl_natbib}

%
%

\end{document}